\documentclass[sigconf,review=false,anonymous=false,authordraft=false]{acmart}
\usepackage{subcaption}
\UseRawInputEncoding
\author{Sunil Aryal}
\affiliation{
  \institution{Deakin University}
  \city{Geelong}
  \state{VIC}
  \country{Australia}
  }
\email{sunil.aryal@deakin.edu.au}

\author{Jonathan R. Wells}
\affiliation{
  \institution{Deakin University}
  \city{Geelong}
  \state{VIC}
  \country{Australia}
  }
\email{j.wells@deakin.edu.au}

\author{Arbind Agrahari Baniya}
\affiliation{
  \institution{Deakin University}
  \city{Geelong}
  \state{VIC}
  \country{Australia}
  }
\email{a.agraharibaniya@deakin.edu.au}

\author{KC Santosh}
\affiliation{
  \institution{University of South Dakota}
  \city{Vermillion}
  \state{South Dakota}
  \country{USA}
  }
\email{Santosh.KC@usd.edu}

\begin{document}

\fancyhead{}

%==== Abstract =====
%\title{Robust preprocessing of data to overcome weaknesses of clustering algorithms}
%\title{Overcoming weaknesses of clustering algorithms by robust preprocessing of data}
\title{Enabling clustering algorithms to detect clusters of varying densities through scale-invariant data preprocessing}

%==== Abstract =====
\begin{abstract}
In this paper, we show that preprocessing data using a variant of rank transformation called `Average Rank over an Ensemble of Sub-samples (ARES)'  makes clustering algorithms robust to data representation and enable them to detect varying density clusters. Our empirical results, obtained using three most widely used clustering algorithms—namely KMeans, DBSCAN, and DP (Density Peak)—across a wide range of real-world datasets, show that clustering after ARES transformation produces better and more consistent results. \\


\end{abstract}

\maketitle

%==== INTRODUCTION =====

\section{Introduction}
\label{sec_intro}
%  2.1 Varying Density Clustering 
%  2.2 ARES 
%  Evaluation  
Clustering is an unsupervised data mining task of partitioning a set of data objects into multiple groups, referred to as `clusters', such that objects within a group are similar to each other and dissimilar to objects in other groups \cite{DataMining_Han2011}. It finds applications in various domains, such as customer/market segmentation in marketing and sales, grouping homologous gene sequences in biology, and detecting communities in social networks. The automatic clustering of real-world data is a challenging task because \cite{DataMining_Han2011}: (i) data can be clustered in several ways based on different perspectives, views or purposes; (ii) real-world data have complex structure with clusters of arbitrary shapes and varying densities; and (iii) in most cases, no prior knowledge about the number and characteristics (\textit{e.g.}, shapes) of clusters is available.

Different types of clustering algorithms have been proposed in the literature. However, there is no single algorithm that is universally effective in detecting various types of clusters across different application domains. Almost all clustering algorithms rely on a notion of (dis)similarity of data. The pairwise (dis)similarity of data objects is estimated using the values of features (also known as attributes) that define these data objects. Since most existing clustering algorithms operate on feature values, they are sensitive to how data features are represented—\textit{i.e.}, their performance may vary significantly if the same data is represented or expressed differently.

As discussed in \cite{SimUSF_Fernando2017,usfAD_Aryal2018,MpDAMI_Aryal2020,usfAD_Aryal2020}, in real-world applications, features of data objects can be captured or collected from various sources/sensors and measured/represented in different forms. For instance, temperature can be measured in $^\circ C$ or $^\circ F$, while sample variability may be expressed as standard deviation ($\sigma$) or variance ($\sigma^2$). Similarly, data can be provided in different scales, such as log likelihood instead of likelihood, or credit risk as income-to-debt ratio or debt-to-income ratio. In the era of internet-of-things, such variation in data representation arises due to: (a) settings of sensors/devices used to measure data; (b) domain/system requirements; and/or (c) data compression to reduce storage and/or transmission costs.    

\begin{figure*}[t]
\centering
\subfloat[$X$]{\includegraphics[width=0.23\textwidth]{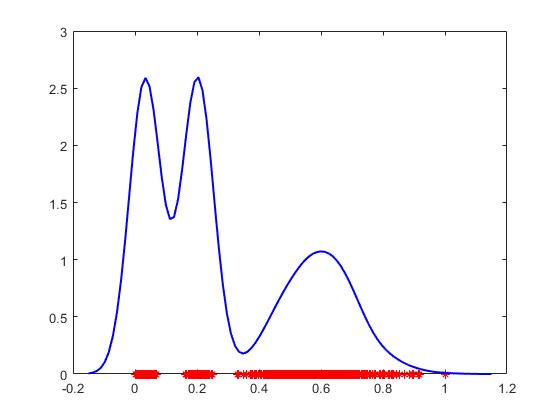}}
\hspace{25pt}
\subfloat[$\log(X)$]{\includegraphics[width=0.23\textwidth]{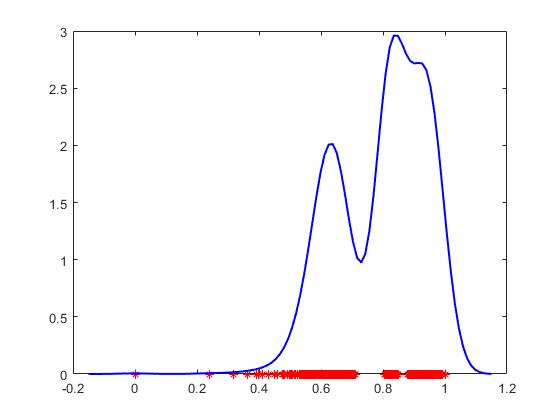}}
\hspace{25pt}
\subfloat[$X^{-1}$]{\includegraphics[width=0.23\textwidth]{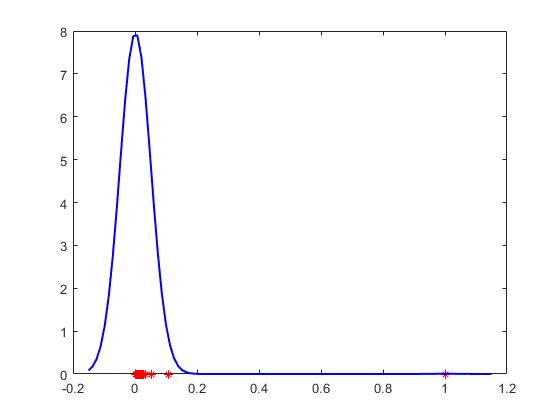}}
\\
\subfloat[Rank($X$)]{\includegraphics[width=0.23\textwidth]{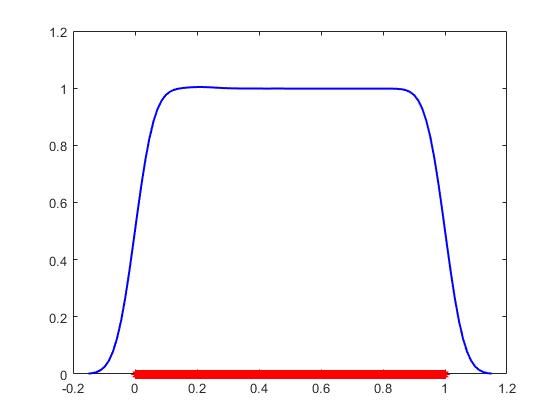}}
\hspace{25pt}
\subfloat[Rank($\log(X)$)]{\includegraphics[width=0.23\textwidth]{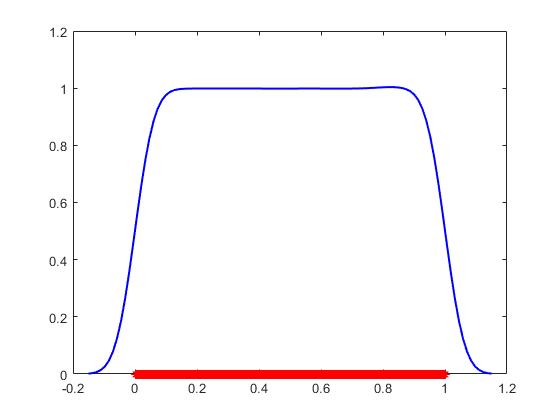}}
\hspace{25pt}
\subfloat[Rank($X^{-1}$)]{\includegraphics[width=0.23\textwidth]{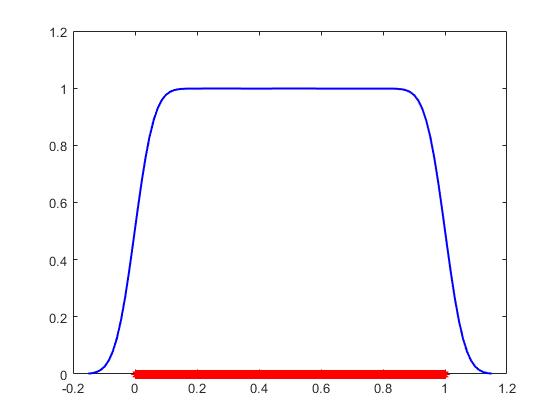}}
\\
\subfloat[ARES($X$)]{\includegraphics[width=0.23\textwidth]{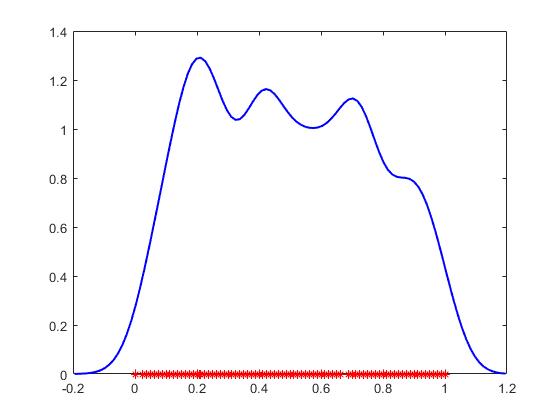}}
\hspace{25pt}
\subfloat[ARES($\log(X)$)]{\includegraphics[width=0.23\textwidth]{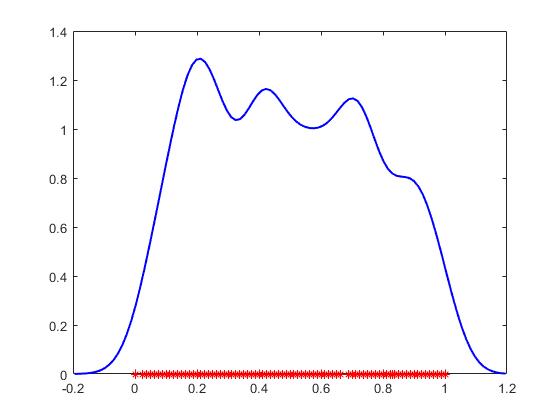}}
\hspace{25pt}
\subfloat[ARES($X^{-1}$)]{\includegraphics[width=0.23\textwidth]{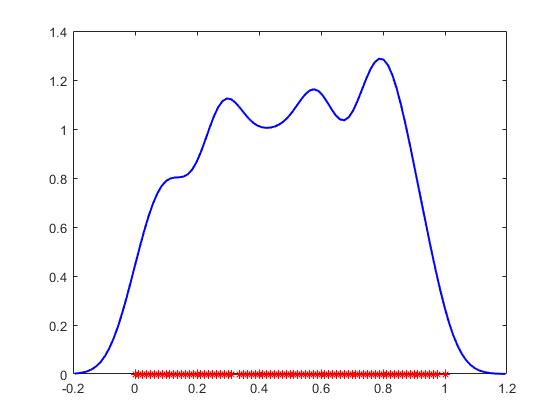}}
\caption{Comparison of data distributions: The first row shows the distributions of an example dataset, along with its logarithmic and inverse scaling. The second and third rows depict the corresponding distributions after the Rank and proposed ARES transformations, respectively. In all cases, data are normalised to be in the range of [0, 1] before modelling the density distribution.}
\label{fig_example_rank_ares}
\end{figure*} 

Different representation of data can be linear or non-linear scaling of each other. Linear scaling poses no problem for pattern recognition, as it does not alter the data distribution. However, non-linear scaling can be problematic, as it can change the distribution of data significantly, thereby providing entirely different insights. It may affect the number, shape, and densities of clusters, making them challenging to detect accurately using existing algorithms. For instance, Fig~\ref{fig_example_rank_ares}a-c show the distributions of a set of one-dimensional data points ($X$) and its logarithmic ($\log(X)$) and inverse ($X^{-1}$) scalings. Clearly, Fig~\ref{fig_example_rank_ares}a has three clusters, with the right most cluster having a lower density compared to the two clusters on the left. However, it changes significantly in Fig~\ref{fig_example_rank_ares}b-c, especially in the case of inverse scaling. Many existing algorithms yield different clustering results for these three representations of the same data. Moreover, some clustering algorithms, particularly those based on density, fail to correctly detect three clusters even in the case of Fig~\ref{fig_example_rank_ares}a because there is no single density threshold $\delta$ to find the three density peaks. They either miss the right cluster with low density (with $\delta>1.2$) or merge the two high density clusters on the left into one cluster (if $\delta<1.3$). The issue of varying density is closely related to the problem of data representation, as non-linear scaling of data alters the density distribution of data.

When given data for clustering, only feature values (numbers) are often provided, and the units/scales in which the data is recorded may remain unknown and unavailable. Even if available, we may not know whether the given representation of data is suitable to detect clusters. We may get misleading insights from data if the given representation is not appropriate. In real-world data clustering, where true clusters are unknown, trying different transformations to find the one yielding the best clustering results is not possible. Even if we can do it, computationally it is not feasible as there are many possible transformations for each attribute resulting in a huge combination to try, particularly in high-dimensional problems. One simple solution is to transform the data in each features into rank (or percentile) space, and using their ranks/percentiles instead of actual data values. Ranks/percentiles are either preserved or reversed even in non-linear scaling of data \cite{SimUSF_Fernando2017}. However, this approach may not be good for clustering, as it tends to make the distribution uniform (see Fig~\ref{fig_example_rank_ares}d-f) because the rank difference between two consecutive values is always 1 regardless of their value differences. It potentially cause some clusters to disappear.

In this paper, we demonstrate that preprocessing data using a variant of rank transformation called `{\bf A}verage {\bf R}ank over an {\bf E}nsemble of {\bf S}ub-samples (ARES)' makes clustering robust to data representation and enabling them to detect clusters with varying densities effectively. The idea is to use rank transformation in multiple subsamples of data and aggregating the ranks. Unlike using rank transformation in the entire dataset, which may mask existing clusters, aggregating ranks form multiple subsamples helps preserve clusters to some extent and remains robust to data representation (see Fig~\ref{fig_example_rank_ares}g-i; note that distribution of $X^{-1}$ is the mirror image of those of $X$ and $\log(X)$ as the ordering of data is reversed). Our empirical results of three most widely used clustering algorithms—KMeans \cite{KMeans_Macqueen1967}, DBSCAN \cite{DBSCAN_Ester1996}, and DP (Density Peak) \cite{DP_Rodriguez2014}—on real-world datasets, demonstrate that clustering after ARES transformation produces better and more consistent results.     

%========= ARES Transformation ===========
\section{ARES Transformation}
\label{sec_relatedWork}

To address the issue of rank transformation resulting in uniform data distribution, we propose a new variant of rank transformation using an ensemble approach. In each feature or dimension $i$, instead of computing rank of $x_i$ among all $n$ values, we propose to aggregate ranks of $x_i$ in $t$ sub-samples of values in dimension $i$. We refer to this proposed method as the ARES ({\bf A}verage {\bf R}ank over an {\bf E}nsemble of {\bf S}ub-samples) transformation.

To simplify the explanation, we assume that $D$ is a one-dimensional dataset where $|D|=n$. We utilise $t$ sub-samples $D_j\subset D$ ($j=1,2,\cdots,t$), where $|D_j|=\psi\ll n$. The transformed value of $x$, denoted as $\tilde{x}_{ARES}$, is the average rank of $x$ across all $t$ sub-samples.
\begin{equation}
    \tilde{x}_{ARES} = \frac{1}{t} \sum_{j=1}^t r(x|D_j)
    \label{eqn_ares}
\end{equation}
where $r(x|D_j)$ represents the rank of $x$ in $D_j$:
\begin{equation}
    r(x|D_j) = |\{y\in D_j: y<x\}|
    \label{eqn_subsamplesRank}
\end{equation}

Based on the definition of $r(x|D_j)$ given by Eqn~\ref{eqn_subsamplesRank}, all $x\in D$ have ranks in the range $\{0,1,\cdots,\psi\}$ in $D_j$. If $s_j^{(1)},s_j^{(2)},\cdots,s_j^{(\psi)}$ are the sorted samples in $D_j$:
  \begin{equation}
    r(x|D_j) =
    \begin{cases}
      0 & \text{if}\ x<s_j^{(1)} \\
      k & \text{if}\ s_j^{(k)}\leq x<s_j^{(k+1)} \text{and } 1\leq k \leq \psi-1\\
      \psi & \text{if}\ x\geq s_j^{(\psi)}
    \end{cases}
  \end{equation}

\begin{figure}[t]
\centering
\subfloat[A given dataset ($D$)]{\includegraphics[width=0.40\textwidth]{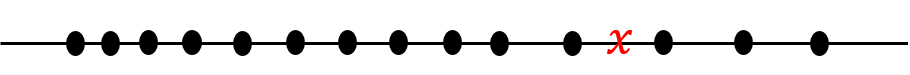}}
\\
\subfloat[$\mathcal{R}_j \mbox{ }(j=1,2,\cdots,t)$]{\includegraphics[width=0.40\textwidth]{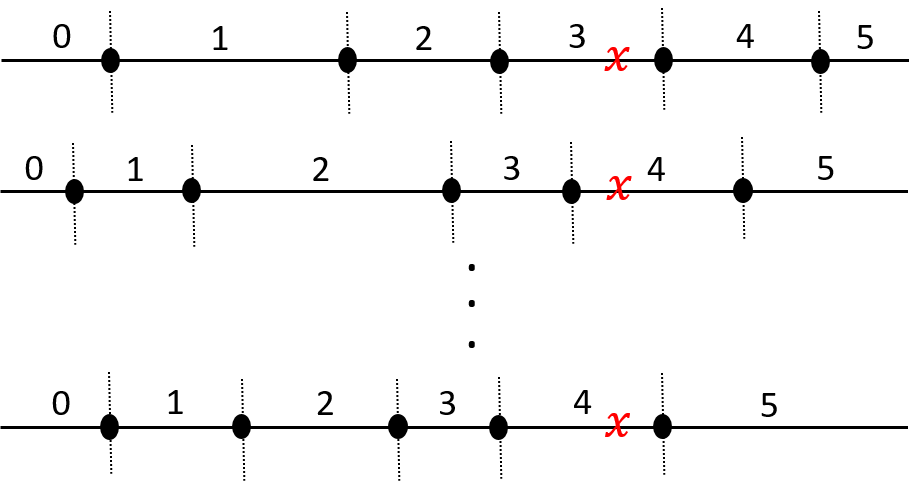}}
\caption{An illustration of ARES: (a) an example of a given dataset $D$; (b) an ensemble of ranking models $\mathcal{R}_j (j=1,2,\cdots,t)$ from sub-samples $D_j\subset D$ and $|D_j|=\psi=5$. Each $\mathcal{R}_j$ is constructed from $D_j$ by partitioning the real domain into $(\psi+1)$ bins which are then ranked as $(0,1,\cdots,\psi)$ from left to right. The ARES transformation of a query point $x$ on red is computed by aggregating its rank in each $\mathcal{R}_j$.}
\label{fig_example_subsamples}
\end{figure} 

For all $x\in D$ such that $s_j^{(k)}\leq x<s_j^{(k+1)}$, they will receive the same $r(x|D_j)$ of $k$, regardless of the their differences in magnitudes. Consequently, they cannot be differentiated. However, it is important to note they may have different ranks in other sub-samples. Therefore, by averaging over various sub-samples, we can maintain their differences to some extent. For instance, even the case of $|D_j|=\psi=1$ where all $x\in D$ have ranks either 0 or 1, depending upon whether they lie to the left or right of the sample selected in $D_j$, repeating this process multiple times (say $t=10$) preserves the differences between data points to some extent. 

Fig~\ref{fig_example_rank_ares}d-i show the distribution of an example dataset and its transformations using traditional rank and ARES. It is evident that ARES better preserves the differences in data values in the original distribution compared to the rank transformation using the entire $D$. Like traditional rank transformation, ARES is robust to changes in scales and units of measurement. ARES, being a variant of rank transformation using sub-samples in an ensemble, it results in the same transformed distribution regardless of the original scaling of data, as shown in Fig~\ref{fig_example_rank_ares}g-i. It is worth noting that the resulting distribution in the case of inverse scale (Fig~\ref{fig_example_rank_ares}i) is the reverse of that in the original and logarithmic scales (Fig~\ref{fig_example_rank_ares}g-h). There is some small differences in the ARES transformations of samples selected in $D_j$ (i.e., $x=s_j^{(k)}$) in the case of inverse scale because of the `$<$' sign in Eqn~\ref{eqn_subsamplesRank}.   

In terms of run-time, generating and sorting $\psi$ samples $t$ times in ARES requires $O(t\psi\log\psi)$ time, and computing ranks for all $n$ instances in $D$ takes $O(nt\log\psi)$ time. On the other hand, the traditional rank transformation involves $O(n\log n)$ time for sorting and ranking the $n$ instances in $D$.

%==== EMPIRICAL RESULTS =====

\section{Empirical Evaluation}
\label{sec_exp}

In this section, we present the clustering results obtained from three most popular clustering algorithms: KMeans \cite{KMeans_Macqueen1967}, DBSCAN \cite{DBSCAN_Ester1996}, and Density Peak (DP) \cite{DP_Rodriguez2014}. We evaluated clustering performances in terms of the quality of clusters produced and robustness to the change in data representation.

\subsection{Experimental setup}

We used seven widely-used real-world datasets to evaluate the clustering results. The characteristics of these datasets are provided in Table~\ref{tbl:datasets}.  

\begin{table}[t]
 \centering
 \caption{Data sets}
 \begin{tabular}{lrrr}
  \toprule
Datasets & \#Instances & \#Attributes & \#Clusters \\
  \midrule
Letters & 20,000 & 16 & 26 \\
Pendigits & 10,992 & 16 & 10 \\
Spambase & 4,601 & 57 & 2 \\
SatImage & 4,435 & 36 & 6 \\
Segment & 2,310 & 16 & 7 \\
Hba & 1,500 & 187 & 15 \\
Gtzan & 1,000 & 230 & 10 \\
%Jain & 373 & 2 & 2 \\
  \bottomrule
 \end{tabular}
\label{tbl:datasets}
\end{table}

\begin{table*}
 \centering
\caption{Clustering results in terms of F1-measure. Best result on each dataset for each clustering algorithm is bold-faced.}
 \begin{tabular}{lcrrrcrrrcrrr}
  \toprule
 & & \multicolumn{3}{c}{DBSCAN} & & \multicolumn{3}{c}{DP} & & \multicolumn{3}{c}{KMeans} \\
  \cmidrule{3-5}
  \cmidrule{7-9}
  \cmidrule{11-13}
Datasets & & min-max & ARES & Rank & & min-max & ARES & Rank & & min-max & ARES & Rank\\
  \midrule
Letters & & 0.2902 & {\bf 0.3271} & 0.2868 & & 0.2923 & {\bf 0.3760} & 0.3310 & & 0.2778 & 0.2905 & {\bf 0.2948} \\
Pendigits & & {\bf 0.6904} & 0.6705 & 0.6860 & & {\bf 0.7763} & 0.7256 & 0.6992 & & 0.6477 & {\bf 0.6933} & 0.6346 \\
Spambase & & 0.3782 & {\bf 0.4590} & 0.4408 & & 0.6574 & {\bf 0.8277} & 0.5692 & & 0.3965 & {\bf 0.8195} & 0.8256 \\
SatImage & & 0.4490 & {\bf 0.4497} & 0.3716 & & 0.6202 & {\bf 0.7235} & 0.6816 & & 0.6431 & {\bf 0.6519} & 0.6509 \\
Segment & & {\bf 0.6702} & 0.6556 & 0.6284 & & {\bf 0.7619} & 0.7367 & 0.6557 & & {\bf 0.5735} & 0.5715 & 0.5472 \\
Hba & & 0.0470 & {\bf 0.1069} & 0.0866 & & 0.2124 & {\bf 0.2444} & 0.1939 & & 0.3246 & {\bf 0.3809} & 0.3700 \\
Gtzan & & 0.0907 & {\bf 0.1021} & 0.0956 & & 0.2479 & {\bf 0.2856} & 0.2597 & & 0.3291 & {\bf 0.3953} & 0.3811 \\
  \bottomrule
 \end{tabular}
\label{tbl:F1-Real}
\end{table*}

We used JAVA implementations of KMeans and DBSCAN available in the WEKA platform \cite{Weka_2009}. We implemented the proposed ARES preprocessing and DP algorithm also in the WEKA platform. There are a number of parameters in these algorithms that need to be set properly for optimal clustering results. We experimented with parameter settings suggested in the respective papers and reported the best results. The number of clusters in DP and KMeans was set to the true number of clusters in the datasets. The parameter $\epsilon$ for DBSCAN and DP was searched in the range of [0.01 to 0.5 with a step size of 0.01] and $minPts$ for DBSCAN in [4, 5, 6, 7, 8]. Similarly, two parameters in ARES were searched as $\psi\in\{2^i|i=0,1,\cdots,5\}$ and $t\in\{10, 25, 50, 100\}$. We used F1-measure as the metric to evaluate cluster results \cite{clusteringMetric_Elke2012}. 

To use existing clustering algorithms, data must be normalised to be in the same range in each dimension. We employed the widely used min-max normalisation technique to ensure data in all dimensions fall within the range of [0,1]. It is interesting to note that such normalisation is not required if ARES or Rank transformation of data is done. In fact, these methods serve as alternatives to traditional preprocessing techniques such as normalisation and standardisation. Unlike normalisation and standardisation, which are sensitive to changes in data representation, ARES and Rank transformations are robust to such changes.   

To assess the robustness of clustering algorithms, we transformed the given data values of each feature $x$ using non-linear scaling methods, including $x^2$, $\sqrt{x}$, $\log{x}$ and $x^{-1}$. Given that the transformations $\log x$ and $1/x$ are undefined for $x = 0$, we applied the transformations to $c(x + \alpha)$, where $\alpha = 0.0001$ and $c = 100$, as done in \cite{MpDAMI_Aryal2020,usfAD_Aryal2020}. Subsequently, the scaled data was normalised within the range of [0,1].

\subsection{Results in the given representation of data}

The clustering results of DBSCAN, DP and KMeans on the given datasets using the default representation of $x$ with min-max, rank, and ARES preprocessing are presented in Table~\ref{tbl:F1-Real}. Our results can be summarised as follows:
\begin{itemize}
    \item The ARES transformation consistently outperformed min-max normalisation and rank transformation across all three clustering algorithms. ARES produced the best results in five datasets for each algorithm. In contrast, with the commonly used min-max normalisation, DBSCAN and DP achieved the best results in two datasets, and KMeans yielded the best result in only one dataset. Rank transformation, on the other hand, resulted in the best outcome for KMeans in the Letters dataset only.
    \item DP consistently outperformed DBSCAN and KMeans across all three preprocessing approaches, aligning with its reputation as a state-of-the-art clustering algorithm. ARES transformation notably improved DP's performance in five datasets, including significant enhancements in cases like Spambase and SatImage. However, there was a slight decrease in performance in two datasets (Pendigits and Segment). 
    \item Rank transformation generally produced worse results than ARES, except with KMeans in Letters, where it produced marginally better results than ARES. Notably, Rank transformation surprisingly produced competitive results comparable to min-max normalisation. This results suggest that while Rank transformation may mask the variation in density distribution in each dimension (see Fig~\ref{fig_example_rank_ares}), it manages to preserve it to some extent in the multidimensional space. 
\end{itemize}

\subsection{Results with different representations of data} 

\begin{table}
 \centering
\caption{F1-measure of DP clustering with different non-linear scalings of data. Best result in each dataset is bold-faced.}
 \begin{tabular}{lrrrrr}
  \toprule
Datasets & $x$ & $x^2$ & $\sqrt{x}$ & $\log x$ & $x^{-1}$ \\
  \midrule
Letters & 0.2923 & {\bf 0.2933} & 0.3054 & 0.2113 & 0.0819 \\
Pendigits & {\bf 0.7763} & 0.6698 & 0.6883 & 0.5469 & 0.4220 \\
Spambase & 0.6574 & 0.3857 & 0.7615 & 0.7223 & {\bf 0.7826} \\
SatImage & {\bf 0.6202} & 0.5018 & 0.5581 & 0.3888 & 0.0706 \\
Segment & 0.7619 & {\bf 0.8065} & 0.7289 & 0.5515 & 0.1694 \\
Hba & 0.2124 & 0.2037 & {\bf 0.2203} & 0.1547 & 0.0241 \\
Gtzan & 0.2479 & 0.2901 & 0.2307 & {\bf 0.3026} & 0.0322 \\
  \bottomrule
 \end{tabular}
\label{tab:dp_scaling}
\end{table}

\begin{figure*}
 \centering
 \subfloat[DP with $x$]{\includegraphics[width=0.23\textwidth]{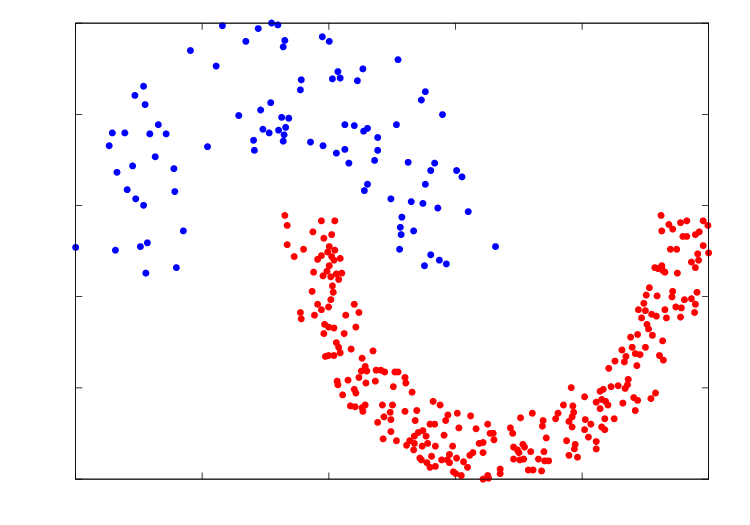}}
 \hspace{25pt}
 \subfloat[DP with $\log x$]{\includegraphics[width=0.23\textwidth]{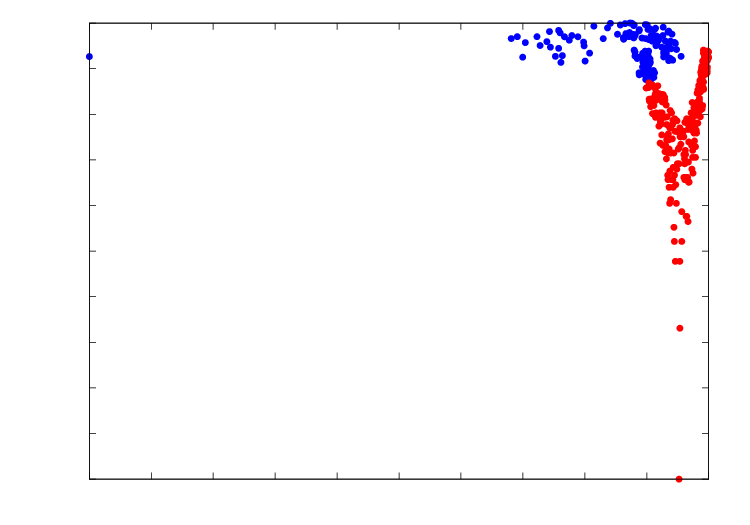}}
 \hspace{25pt}
 \subfloat[DP with $x^{-1}$]{\includegraphics[width=0.23\textwidth]{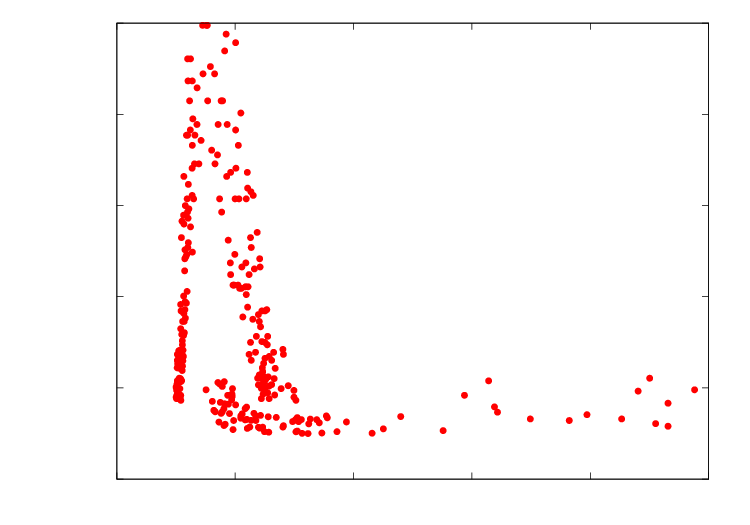}} \\
 \subfloat[DP with ARES($x$)]{\includegraphics[width=0.23\textwidth]{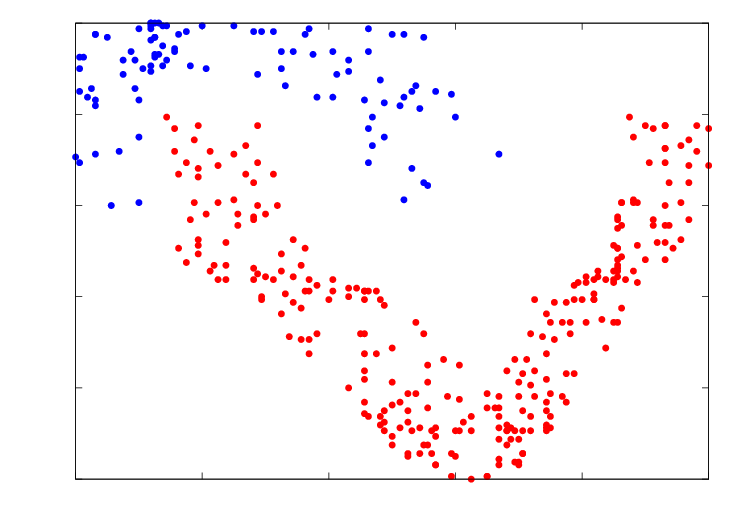}}
 \hspace{25pt}
 \subfloat[DP with ARES($\log x$)]{\includegraphics[width=0.23\textwidth]{./jain_dp_ares_x}}
 \hspace{25pt}
 \subfloat[DP with ARES($x^{-1}$)]{\includegraphics[width=0.23\textwidth]{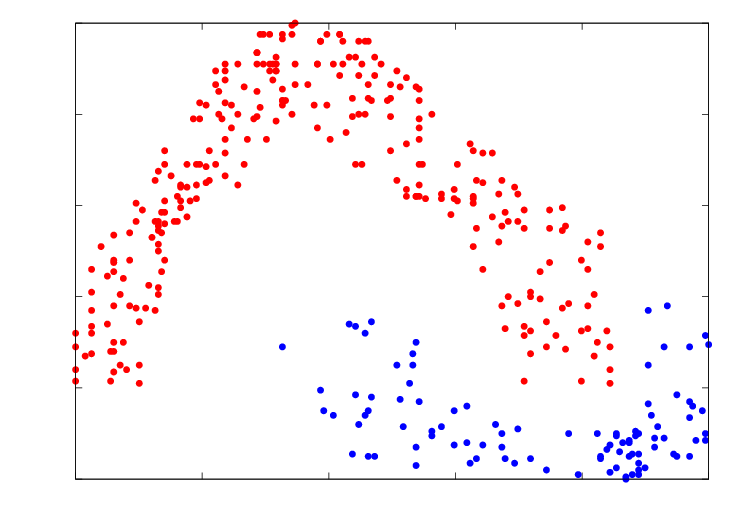}}
 \caption{Clustering results of the Jain dataset with $x$, $\log{x}$, and $x^{-1}$ scaling using DP, both with and without ARES transformation. Note that in the case of DP with $x^{-1}$ (Subfig c), only a portion of the plot is shown to provide a clearer view of most points. Due to some extreme values, displaying all points in this section would result in grouping most of them into a single point. Additionally, another small cluster was detected in the unseen part of the plot.}
 \label{fig_jain}
\end{figure*}

To analyse the robustness of the clustering algorithms to changes in data representation, we applied them after non-liner scaling of data with $x^2$, $\sqrt{x}$, $\log x$ and $x^{-1}$. The clustering results of all three algorithms with min-max normalisation exhibited variations with the changes in data representation. In contrast, ARES and Rank transformation of data consistently produced the same results across all these representations, as shown in Table~\ref{tbl:F1-Real}, with very small differences in the case of $x^{-1}$ scaling in some cases. This behaviour aligns with the discussion presented earlier in Section~\ref{sec_relatedWork}.

The results of the DP clustering algorithm, which performed the best among the three algorithms, with different representations of real-world datasets are presented in Table~\ref{tab:dp_scaling}. Similar trends were observed with DBSCAN and KMeans. It is interesting to note that the given representation of data produced the best results in only two datasets, and no presentation yielded the best results in more than two datasets when compared to ARES (Table~\ref{tbl:F1-Real}). These findings suggest that the default data representation may not be optimal for achieving the best clustering results. Therefore, applying ARES preprocessing is a preferred and reliable option.

Visual examples of DP in  the two-dimensional synthetic dataset called Jain in three representations of $x$, $\log x$ and $x^{-1}$, with and without ARES transformations, are shown in Fig~\ref{fig_jain}. DP without ARES transformation resulted F1-measure of 1.0000, 0.8607 and 0.4241 with $x$, $\log x$ and $x^{-1}$ representations, respectively. On the other hand, DP with ARES transformation consistently produced perfect clustering results, achieving an F1-measure of 1.0000 in all three cases.

%==== CONCLUSIONS ======

\section{Conclusions and Future Work}
\label{sec_con}

In real-world problems, data are collected from various sources and sensors, which can be represented in different forms/formats. The given representation of data may not be appropriate for obtaining a good clustering result using existing algorithms, as their performances are sensitive to data representation. They can give misleading clustering results. It is not practical to try different representations to see which one gives the best result because: (i) there can be infinitely many possible combinations to try in high-dimensional problems; and (ii) there may not be ground truth clustering result available to compare the produced clustering results. Therefore, it is preferred to use algorithms that are robust to data representations (units/scales of measurement). 

In this paper, we proposed one simple approach to enhance robustness of clustering algorithms to the change in data representations. We propose a robust data processing technique called ARES that results in the same transformed distribution regardless of the input data representations. Our results demonstrate that applying existing clustering algorithms after preprocessing data with ARES transformation yields in better, reliable and robust clustering results.

In future work, we plan to evaluate performance of other clustering algorithms with and without ARES transformation. Additionally, exploring the applicability of ARES in other machine learning problems is an interesting avenue for further investigation.

%\begin{figure}[t]
% \centering
% \subfloat[DP with $x$]{\includecombinedgraphics[scale=0.31]{./graphs/jain_dp_x}}
% \subfloat[DP with ARES($x$)]{\includecombinedgraphics[scale=0.31]{./graphs/jain_dp_ares_x}} \\
% \subfloat[DP with $\log x$]{\includecombinedgraphics[scale=0.31]{./graphs/jain_dp_ln_x}}
% \subfloat[DP with ARES($\log x$)]{\includecombinedgraphics[scale=0.31]{./graphs/jain_dp_ares_x}} \\
% \subfloat[DP with $x^{-1}$]{\includecombinedgraphics[scale=0.31]{./graphs/jain_dp_inv_x}}
% \subfloat[DP with ARES($x^{-1}$)]{\includecombinedgraphics[scale=0.31]{./graphs/jain_dp_ares_inv_x}}
% \caption{Clustering results of the Jain dataset with $x$, $\log{x}$, and $x^{-1}$ scaling using DP, both with and without ARES transformation. Note that in the case of DP with $x^{-1}$ (Subfig e), only a portion of the plot is shown to provide a clearer view of most points. Due to some extreme values, displaying all points in this section would result in grouping most of them into a single point. Additionally, another small cluster was detected in the unseen part of the plot.}
% \label{fig_jain}
%\end{figure}

%==== ACKNOWLEDGEMENTS ====

\section*{Acknowledgement}
This research is funded by the US Air Force Office of Scientific Research (AFOSR) under grant number FA2386-20-1-4005.

%==== BIBLIOGRAPHY ======

\bibliographystyle{unsrt}
\bibliography{manuscript}

\begin{thebibliography}{10}

\bibitem{DataMining_Han2011}
Jiawei Han, Micheline Kamber, and Jian Pei.
\newblock {\em Data Mining: Concepts and Techniques}.
\newblock Morgan Kaufmann Publishers, third edition, 2011.

\bibitem{SimUSF_Fernando2017}
Thilak~L. Fernando and Geoffrey~I. Webb.
\newblock {SimUSF: an efficient and effective similarity measure that is invariant to violations of the interval scale assumption}.
\newblock {\em Data Mining and Knowledge Discovery}, 31(1):264--286, 2017.

\bibitem{usfAD_Aryal2018}
Sunil Aryal.
\newblock Anomaly detection technique robust to units and scales of measurement.
\newblock In {\em Proceedings of the 22nd Pacific-Asia Conference on Knowledge Discovery and Data Mining}, pages 589--601, 2018.

\bibitem{MpDAMI_Aryal2020}
Sunil Aryal, Kai~Ming Ting, Takashi Washio, and Gholamreza Haffari.
\newblock A comparative study of data-dependent approaches without learning in measuring similarities of data objects.
\newblock {\em Data mining and knowledge discovery}, 34(1):124--162, 2020.

\bibitem{usfAD_Aryal2020}
Sunil Aryal, K.C. Santosh, and Richard Dazeley.
\newblock {usfAD}: a robust anomaly detector based on unsupervised stochastic forest.
\newblock {\em International Journal of Machine Learning and Cybernetics}, 12:1137--1150, 2021.

\bibitem{KMeans_Macqueen1967}
J.~Macqueen.
\newblock Some methods for classification and analysis of multivariate observations.
\newblock In {\em Proceedings of the Fifth Berkeley Symposium on Mathematical Statistics and Probability}, pages 281--297, 1967.

\bibitem{DBSCAN_Ester1996}
Martin Ester, Hans-Peter Kriegel, Joerg Sander, and Xiaowei Xu.
\newblock A density-based algorithm for discovering clusters in large spatial databases with noise.
\newblock In {\em Proceedings of the ACM SIGKDD}, pages 226--231, 1996.

\bibitem{DP_Rodriguez2014}
Alex Rodriguez and Alessandro Laio.
\newblock Clustering by fast search and find of density peaks.
\newblock {\em science}, 344(6191):1492--1496, 2014.

\bibitem{Weka_2009}
Mark Hall, Eibe Frank, Geoffrey Holmes, Bernhard Pfahringer, Peter Reutemann, and Ian~H. Witten.
\newblock The weka data mining software: An update.
\newblock {\em SIGKDD Exploration Newsletter}, 11(1):10--18, 2009.

\bibitem{clusteringMetric_Elke2012}
Elke Achtert, Sascha Goldhofer, Hans-Peter Kriegel, Erich Schubert, and Arthur Zimek.
\newblock Evaluation of clusterings -- metrics and visual support.
\newblock In {\em Proceedings of the 2012 IEEE 28th International Conference on Data Engineering}, pages 1285--1288, 2012.

\end{thebibliography}

\end{document}